\PassOptionsToPackage{dvipsnames, svgnames, x11names}{xcolor}
\documentclass[journal]{IEEEtran}
\usepackage{amsmath,amsfonts}
\usepackage{algorithmic}
\usepackage{algorithm}
\usepackage{array}
\usepackage[caption=false,font=scriptsize,labelfont=rm,textfont=rm]{subfig}
\usepackage{textcomp}
\usepackage{stfloats}
\usepackage{url}
\usepackage{verbatim}
\usepackage{adjustbox}
\usepackage{multirow} 
\usepackage{booktabs} 
\usepackage{adjustbox}
\usepackage{amssymb}     
\usepackage{subcaption}
\usepackage{enumitem}
\usepackage[percent]{overpic}
\usepackage{graphics}
\usepackage{caption}
\usepackage{adjustbox}
\usepackage{epstopdf}
\usepackage{CJKutf8}
\usepackage{orcidlink} 
\usepackage{colortbl}
\usepackage[numbers, sort&compress]{natbib}
\usepackage{graphicx} 

\begin{document}
\title{PillarMamba: Learning Local-Global Context for Roadside Point Cloud via Hybrid State Space Model}
\author{Zhang Zhang$^{\orcidlink{0009-0000-7253-4018}}$, Chao Sun$^{\orcidlink{0000-0002-9324-0892}}$, Chao Yue, Da Wen, Tianze Wang, Jianghao Leng
        
\thanks{This work was supported by Key-Area Research and Development Program of Guangdong Province (2023B0909040001). (Corresponding author: Chao Sun.)}
\thanks{Zhang Zhang, Chao Sun, Chao Yue, Da Wen, Tianze Wang and Jianghao Leng are with the National Engineering Laboratory for Electric Vehicles, School of Mechanical Engineering, Beijing Institute of Technology, Beijing 100081, China (email:  zhangzhang00@bit.edu.cn, chaosun@bit.edu.cn, 3120235513@bit.edu.cn, wenda316@163.com, 3120230483@bit.edu.cn, 3120195252@bit.edu.cn}
\thanks{Manuscript received April 19, 2021; revised August 16, 2021.}}

\markboth{Journal of \LaTeX\ Class Files,~Vol.~14, No.~8, August~2021}%
{Shell \MakeLowercase{\textit{et al.}}: A Sample Article Using IEEEtran.cls for IEEE Journals}
\IEEEpubid{0000--0000/00\$00.00~\copyright~2021 IEEE}
\maketitle

\begin{abstract}
Serving the Intelligent Transport System (ITS) and Vehicle-to-Everything (V2X) tasks, roadside perception has received increasing attention in recent years, as it can extend the perception range of connected vehicles and improve traffic safety. However, roadside point cloud oriented 3D object detection has not been effectively explored. To some extent, the key to the performance of a point cloud detector lies in the receptive field of the network and the ability to effectively utilize the scene context. The recent emergence of Mamba, based on State Space Model (SSM), has shaken up the traditional convolution and transformers that have long been the foundational building blocks, due to its efficient global receptive field. In this work, we introduce Mamba to pillar-based roadside point cloud perception and propose a framework based on Cross-stage State-space Group (CSG), called PillarMamba. It enhances the expressiveness of the network and achieves efficient computation through cross-stage feature fusion. However, due to the limitations of scan directions, state space model faces local connection disrupted and historical relationship forgotten. To address this, we propose the Hybrid State-space Block (HSB) to obtain the local-global context of roadside point cloud. Specifically, it enhances neighborhood connections through local convolution and preserves historical memory through residual attention. The proposed method outperforms the state-of-the-art methods on the popular large scale roadside benchmark: DAIR-V2X-I. The code will be released soon.
\end{abstract}
\begin{IEEEkeywords}
point cloud, roadside, object detection.
\end{IEEEkeywords}

\section{Introduction}
\IEEEPARstart{R}{oadside} perception, as an important component of Intelligent Transport System (ITS), aims to well complement the environment estimation of connected vehicles due to its natural perspective advantage. With the recent release of large-scale real-world benchmarks \cite{dair}, \cite{rope} for roadside scenes, the state-of-the-art performance \cite{heightformer}, \cite{bevspread}, {\cite{bevheight} has been constantly refreshed recently. However, roadside point cloud oriented 3D object detection has not been effectively explored. 3D object detection in roadside point cloud plays a crucial role in roadside perception, which has accurate distance information relative to visual-only perception and helps the system to accurately estimate the surrounding environment. Due to the lack of effective exploration and specific design of roadside-centric point cloud perception, it is difficult for current roadside perception methods to meet the needs of ITS. The current mainstream vehicle-side methods \cite{pointrcnn}, \cite{pvcnn}, \cite{pvrcnn}, \cite{second}, \cite{pillarnet}, \cite{pillarnext} are based on sparse backbone networks and have made significant advances by relying on the sparsity of vehicle-side point cloud. Due to the different installation locations, the perspective of roadside sensors are significantly different from that of the vehicle side, which results in a denser coverage compared to the vehicle-side point cloud in the Bird's Eye View (BEV). Vehicle-side methods cannot achieve outstanding performance by simply migrating to roadside scenarios.

To some extent, the key to the performance of a roadside point cloud detector lies in the receptive field of the network \cite{pillarnet}, \cite{pillarnext}, \cite{aspp} and the ability to effectively utilize scene context. A large receptive field with dense context allows the network to capture spatial cues from a wider area, enabling more information to be shared between multi-objects and creating a remote spatial connection. Recently, State Space Model (SSM) \cite{mamba}, \cite{vmamba}, \cite{mambair}, \cite{qmambair} have been proposed to model long-range dependencies and dense feature extraction, achieving significant results efficiently. Specifically, with the discretized state space equations in recursive form and the parallel scan algorithm, Mamba can efficiently model long-range dependencies. The above promising properties motivate us to explore the potential of Mamba in roadside point cloud scenarios with dense context for efficient point cloud 3D object detection. Therefore, we introduce Mamba to pillar-based roadside point cloud perception. However, in subsequent experiments we found that it still faces challenges.

On the one hand, recent roadside point cloud detectors utilize their dense backbone networks with high resolution to obtain a rich context of the scene, which also imposes a high computational burden. Although Mamba relies on its linear attention mechanism \cite{demystify}, \cite{mamba} to reduce computational overhead and achieve performance acceleration through parallel operations, it still faces unacceptably high overhead in a roadside point cloud detector with dense context. Therefore, we propose the Cross-stage State-space Group (CSG), which efficiently extracts the global context of roadside point cloud through cross-stage connections and the extended receptive field. 
\IEEEpubidadjcol
Specifically, it achieves computational burden through channel dimension reduction, channel split, and channel connections. It not only achieves computational efficiency and thus inference acceleration, but also allows different layers to complement each other through cross stage connections and enhances the expressive ability of networks.

On the other hand, whether it is the standard Mamba \cite{mamba} designed for 1D sequences in Natural Language Processing (NLP) or visual Mamba \cite{vmamba} designed for 2D image sequences in Computer Vision (CV), it is limited by its relying on scanning flattened sequences in a recursive manner. We find that this recursive scanning mode faces two main problems in roadside point cloud: local connection disrupted and historical relationship forgotten. As shown in Figure \ref{fig_1}, we assume a roadside point cloud scene with vehicles in BEV map, where the \textcolor{orange}{orange}, \textcolor{RoyalBlue}{blue} and white grids denote the vehicle A, B and empty. In this scene, vehicle A and B are in close neighborhood in 3D space. When we use visual Mamba to process the BEV map of the current scene, it flattens the grids in the map into 4 \textcolor{ForestGreen}{scans}, called Cross-Scan. 

First, since Cross-Scan deals with flattened 1D sequences in a recursive manner, it may result in spatially similar grids being at very distant locations in the sequence, causing local connection disrupted. It can be seen that the orange and blue grids in the red box are a long way apart in the sequence. The long distances in the sequence disrupt the neighborhood connections in the original 3D space, limiting the network's ability to understand the space. Second, even though the roadside point cloud is denser compared to the vehicle side, it still faces a large number of empty grids. In 1D sequences with a large number of empty grids, it is extremely easy for small objects to be drowned in historical invalid information of recursive equations. It can be seen that the orange and blue grids in the red box face a large number of empty grids both before and after. The drowning out of small objects weakens the network's ability to distinguish between foreground and background.

\begin{figure*}[!t]
\centering
\includegraphics[width=.99\textwidth]{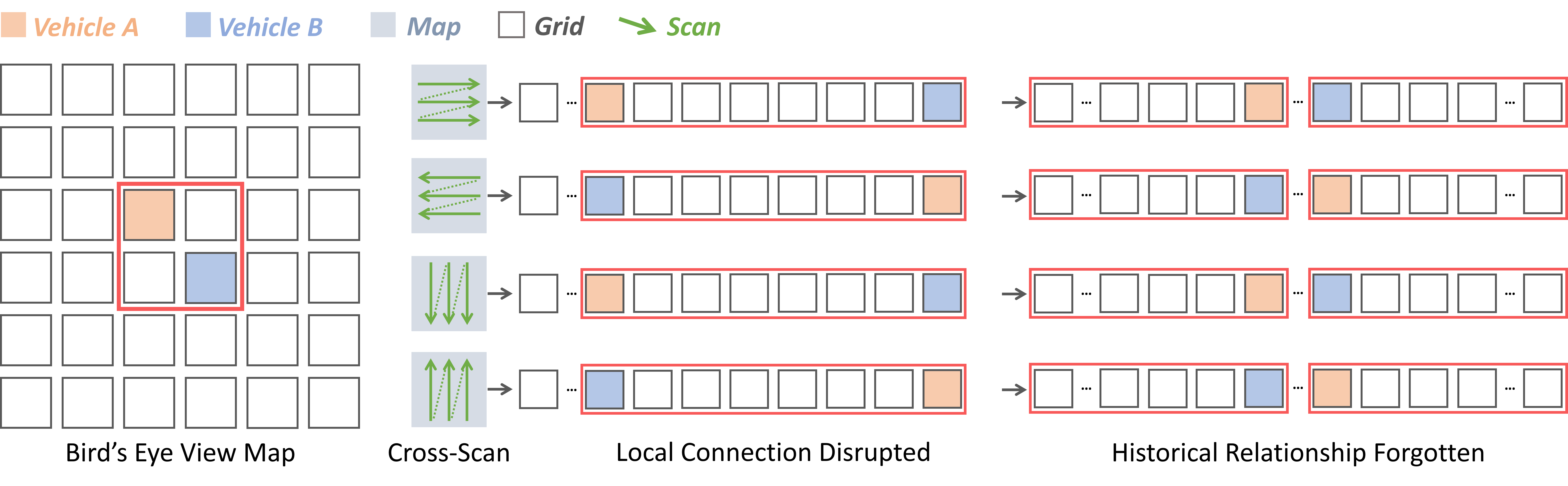}
\caption{We assume a roadside point cloud scene with vehicles in BEV map, where the \textcolor{orange}{orange} and \textcolor{RoyalBlue}{blue} grids denote the vehicle A and B. In this scene, vehicle A and B are in close neighborhood in 3D space. When we use visual Mamba to process the BEV map of the current scene, it divides the grids in the map into 4 \textcolor{ForestGreen}{scans}, called Cross-Scan. The traditional visual Mamba methods simply flatten the tokens of BEV map into 1D sequences by scanning them row-by-row or column-by-column. It not only increases the distance between neighborhood tokens in BEV map, and also small objects to be drowned in historical invalid information of recursive equations, which weakens the local dependencies and blurs the foreground and background.}
\label{fig_1}
\vspace{-0.2cm}
\end{figure*}

To address the above challenges, we introduce the Hybrid State-space Block (HSB) to obtain the local-global context of roadside point cloud. Specifically, it enhances neighborhood connections through local convolution and preserves historical memory through residual attention.

Levaraging on the findings above, we introduce PillarMamba to adapt Mamba for roadside point cloud, which consists of Cross-stage State-space Group (CSG) and Hybrid State-space Block (HSB). The proposed method outperforms the state-of-the-art methods on the popular real-world large-scale roadside benchmark: DAIR-V2X-I \cite{dair}.

Our main contributions are as follows:
\begin{itemize}

\item We explore the application of the state space model to roadside scenarios. Specifically, by utilizing its linearly expanding attention and efficient global receptive field, we effectively extract the spatial connections of objects in roadside point cloud.\\

\item We utilize the Cross-stage State-space Group (CSG) for efficient computation in dense context. It not only achieves computational efficiency and thus inference acceleration, but also allows different layers to complement each other through cross stage connections, which enhances the expressive ability of networks.\\

\item Hybrid State-space Block (HSB) is proposed to boost the power of the standard state space model and address the local connection disrupted and historical relationship forgotten. It enhances the network's spatial understanding and preserves the historical memory in recursive state space equation. 

\end{itemize}

\section{Related Work}

\textbf{Vehicle-side Point Cloud 3D Object Detection.}
The development of point cloud object detection has brought profound impact on autonomous driving. PointRCNN \cite{pointrcnn} generates high-quality 3D proposals directly from the point cloud to subsequently obtains the box refinement and confidence prediction. 3DSSD \cite{3dssd} achieves decent performance with downsample strategy and candidate point generation network. Above point-based methods naturally preserve the exact positional relationships of the point cloud in 3D space, enabling local feature aggregation. VoxelNet \cite{voxelnet} splits the point cloud into 3D voxels and converts points in each voxel into a uniform feature representation via voxel feature encoding layers. SECOND \cite{second} improves efficiency and performance by introducing sparse 3D convolution. PointPillars \cite{pointpillars} learns to represent point cloud organised in pillar representation, which utilises dense 2D convolution to achieve efficient single-stage 3D object detector. Recent methods \cite{pillarnet}, \cite{pillarnext}, \cite{sst} use a sparse backbone network based on pillar representation instead of dense to fit the sparse point cloud on the vehicle side. They achieve significant results by expanding the BEV receptive field of the network through effective design, benefiting from the irregularity and sparsity of the vehicle-side point cloud scenes.

\textbf{Roadside Point Cloud 3D Object Detection.}
Point cloud 3D object detection in autonomous driving is currently limited to ego vehicles \cite{kitti}, \cite{waymo}, \cite{nuscenes}, \cite{argo}, \cite{argo2}. Roadside point cloud 3D object detection as part of ITS has undeniable advantages. It has a relatively long perception range and extends the perception range beyond the limitations of ego vehicles, enhancing the road safety. However, roadside point cloud 3D object detection is largely unexplored. Recently, some large-scale real-world datasets towards roadside scenes have been published for facilitating the development of roadside 3D object detection tasks \cite{dair}, \cite{v2xseq}, \cite{rope}. Unlike the point cloud captured by vehicle-side sensors, the roadside point cloud has a denser coverage in BEV map and richer dense context.

\textbf{State Space Model.}
State Space Model \cite{ssm}, derived from classical control theory, has recently been introduced to deep learning \cite{s4}, \cite{s5}, gaining increasing attention. Mamba \cite{mamba} is a data-dependent SSM with a selection mechanism that outperforms Transformers in natural language. Its computational complexity scales linearly with the length of the input sequence and yields an efficient global receptive field based on an effective hardware design and parallel computation. Since the introduction of Mamba, a number of efforts have been proposed to leverage its capability for vision applications.  Vim \cite{vim} introduces a bi-directional SSM scheme that processes tokens in the forward and backward directions to capture global context and improve spatial comprehension. VMamba \cite{vmamba} proposes a Cross-Scan module. This module uses a four-way selective scan methodology to integrate information from surrounding tokens and capture the global context. Despite the significant advances in vision made by Mamba as described above, it still faces challenges in roadside point cloud. The SSM-based methods suffer from local connection
disrupted and historical relationship forgotten due to the spatial neighborhood properties in the BEV maps and the large number of empty grids in the sequences.

\begin{figure*}[!t]
\centering
\includegraphics[width=.99\textwidth]{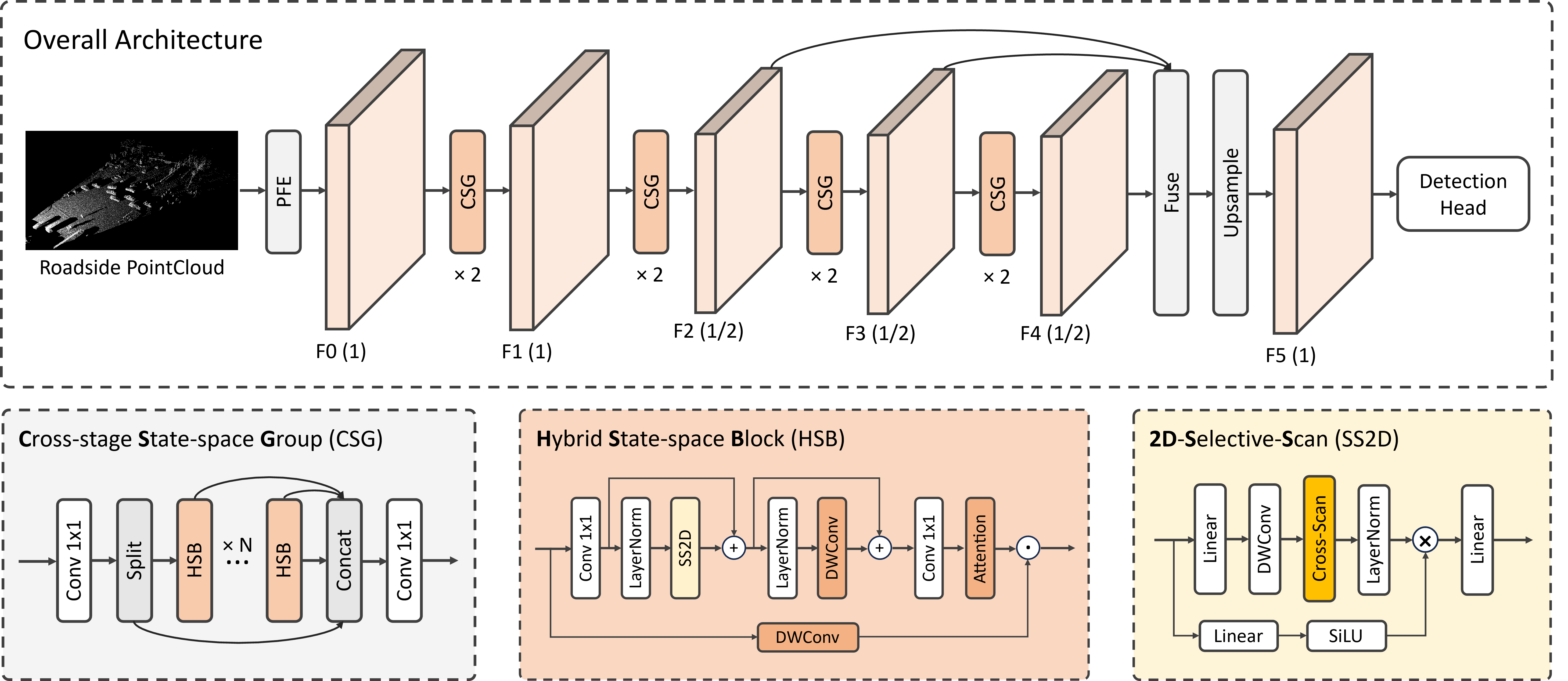}
\caption{The proposed PillarMamba consists of Pillar Feature Encoder (PFE), Backbone and Detection Head. We first convert the roadside point cloud into dense BEV feature map $F_0 \in R^{C\times X\times Y}$ based on PFE. We use Cross-stage State-space Group (CSG) as the basic feature extractor in dense backbone network. $F_1 \in R^{C\times X\times Y}$ is obtained by CSG with 2 HSB layers. Then, $F_1$ is fed into a down-sample layer and CSG with 2 HSB layers, obtain the $F_2 \in R^{C\times \frac{X}{2}\times \frac{Y}{2}}$, which is used to further conserve memory. As above, we obtain $F_3$ and $F_4$ in turn. Then, we concatenate $F_2$, $F_3$ and $F_4$ in channel dimension to integrate multi-scale information. We then feed the features into the up-sample layer to get the $F_5 \in R^{C\times X\times Y}$. Detection Head first encodes the pillar features with a convolutional layer and then predicts the 3D bounding box consisting of locations $(x, y, z)$, dimensions $(l, w, h)$ and yaw angle $\theta$.}
\label{fig_2}
\vspace{-0.2cm}
\end{figure*}

\section{Method}
We first briefly introduce the problem definition of 3D object detection based on roadside point cloud. Then, we describe our framework in detail.

\subsection{Problem Definition}
In this work, we would like to detect the 3D bounding boxes based on the given scene point cloud. We get the point cloud $ P \in R^{N \times 4}$ from the roadside LIDAR with coordinates $x$, $y$, $z$ and reflectance $r$, where $N$ represents the number of points in the point cloud. We would like to detect the 3D bounding boxes of objects $B \in R^{M \times 7} $ from point cloud  with locations $(x, y, z)$, dimensions $(l, w, h)$ and the yaw angle $\theta$, where $M$ represents the number of 3D bounding boxes.

\subsection{Overall Architecture}
The proposed PillarMamba consists of Pillar Feature Encoder (PFE) \cite{pointpillars}, Backbone and Detection Head \cite{centerpoint}, as shown in Figure. \ref{fig_2}. We first convert the roadside point cloud into dense BEV feature map $F_0 \in R^{C\times X\times Y}$ based on PFE. We use Cross-stage State-space Group (CSG) as the basic feature extractor in dense backbone network. $F_1 \in R^{C\times X\times Y}$ is obtained by CSG with 2 HSB layers. Then, $F_1$ is fed into a down-sample layer and CSG with 2 HSB layers, obtain the $F_2 \in R^{C\times \frac{X}{2}\times \frac{Y}{2}}$, which is used to further conserve memory. As above, we obtain $F_3$ and $F_4$ in turn. Then, we concatenate $F_2$, $F_3$ and $F_4$ in channel dimension to integrate multi-scale information. We then feed the features into the up-sample layer to get the $F_5 \in R^{C\times X\times Y}$. Detection Head first encodes the pillar features with a convolutional layer and then predicts the 3D bounding box consisting of locations $(x, y, z)$, dimensions $(l, w, h)$ and yaw angle $\theta$.

\subsection{Pillar Feature Encoder}
We follow the previous work \cite{pointpillars} to encode the roadside point cloud. We extend the new coordinates by calculating the average coordinates of all points in the pillar and their offsets with respect to the center point. The original 4-dimensional coordinates are expanded to 9-dimensional coordinates, resulting in a point cloud with richer geometric information. Then, we converted the discrete point cloud into a dense BEV feature map of size $(C, X, Y)$ to facilitate the application of efficient 2D dense convolution, where $X$, $Y$ and $C$ indicate the x-direction distance, y-direction distance of the BEV space and the feature channel dimension, respectively.

\subsection{Cross-stage State-space Group}
Although Mamba relies on its linear attention mechanism \cite{demystify}, \cite{mamba} to reduce computational overhead and achieve performance acceleration through parallel operations, it still faces unacceptably high overhead in a roadside point cloud detector with dense context. Therefore, we propose the Cross-stage State-space Group (CSG), which efficiently extracts the global context of roadside point cloud through cross-stage connections and the extended receptive field, as shown in Figure \ref{fig_2}. The input feature map $F$ is processed to conserve memory as follows:
\begin{equation}
    F_{d1}, F_{d2} = Split(Conv_{down}(F))
    \label{eq1}
\end{equation}

Where $Conv_{down}$ and $Split$ denote the convolution layer for channel dimension down-sample with kernel size 1 and split along the channel dimension of the feature map. We then perform feature extraction and recover the feature map by concatenation and convolution.
\begin{equation}
    F_{out} = Conv_{up}(Concat(HSB(F_{d1}), F_{d2}))
    \label{eq2}
\end{equation}

Where $Conv_{up}$ and $Concat$ denote the convolution layer for channel dimension up-sample with kernel size 1 and concatenation operation along the channel dimension for feature map. And $HSB$ can be replaced by other feature extraction operators. 

With the proposed CSG, we achieve a trade-off between performance and computational burden based on state space model.

\subsection{Hybrid State-space Block}
The promising efficient global receptive field and effective extraction of dense context motivate us to explore the potential of SSM in roadside point cloud scenarios. We introduce SSM to pillar-based roadside point cloud perception, however, it still faces the the local connection disrupted and historical relationship forgotten, which weakens the network's spatial understanding and limits the network's performance in small object categories. Therefore, we propose the simple but effective Hybrid State-space Block (HSB) to obtain the local-global context of roadside point cloud, as shown in Figure \ref{fig_2}. The feature map first goes through the original SS2D block \cite{vmamba} as shown in below:
\begin{equation}
\begin{aligned}
    F_{down} &= Conv_{down}(F)\\
    F_{down} &= SS2D(LayerNorm(F_{down}) + F_{down}
\end{aligned}
\label{eq3}
\end{equation}

We utilize SS2D block based on state-space model to obtain an efficient global receptive field. We then replaced the original MLP layer with a local deep-wise convolution (e.g., kernel size 3) to perform local feature enhancement and maintain local neighborhood connections, as shown in below:
\begin{equation}
\begin{aligned}
    F_{down} &= DWConv(LayerNorm(F_{down}) + F_{down}\\
    F_{up} &= Conv_{up}(F_{down})
\end{aligned}
\label{eq4}
\end{equation}

Then we preserve the historical memory through residual attention: 
\begin{equation}
    F_{out} = Attention(F_{up}) \cdot DWConv(F)
    \label{eq5}
\end{equation}

Where $Attention$ denotes the operator that imposes weights on channel dimension of feature map based on global average pooling following SE block \cite{se}. 

With the proposed HSB, we enhance the neighborhood connections and preserve the historical memory.

\subsection{2D-Selective-Scan}
The core mechanism of SS2D block is Cross-Scan module. This module uses a four-way selective scan methodology to integrate information from surrounding tokens and capture the global context. For each scan, it flattens the BEV map into a 1D sequence and uses recursive state space equations to obtain long-range dependencies. A 1D continuous input $x(t) \in \mathbb{R}$ is transformed into $y(t) \in \mathbb{R}$ via a learnable hidden state $h(t) \in \mathbb{R}^{M}$ with parameters $A \in \mathbb{R}^{M\times M}$, $B \in \mathbb{R}^{M\times 1}$ and $C \in \mathbb{R}^{1\times M}$ according to:
\begin{equation}
\begin{aligned}
    h'(t) &= A\cdot h(t) + B \cdot x(t)\\
    y(t) &= C \cdot h(t)
\end{aligned}
\label{eq6}
\end{equation}

In order to improve the computational efficiency, the continuous parameters $A$, $B$ and $C$ in the above formulation are further converted into discrete parameters \cite{gu2021combining}. Specifically, assuming the time scale $\Delta$, the discrete parameters $\overline{A} \in \mathbb{R}^{M\times M}$, $\overline{B} \in \mathbb{R}^{M\times 1}$ and $\overline{C} \in \mathbb{R}^{1\times M}$ can be obtained by applying the zero-order hold rule:
\begin{equation}
\begin{aligned}
    \overline{A} &= {\rm exp}(\Delta A)\\
    \overline{B} &= (\Delta A)^{-1} \cdot ({\rm exp}(\Delta A) -I) \cdot (\Delta B) \\
    \overline{C} &= C
\end{aligned}
\label{eq7}
\end{equation}

Then, the Equation \eqref{eq6} can be expressed with discrete parameters as:
\begin{equation}
\begin{aligned}
    h(t) &= \overline{A}\cdot h(t-1) + \overline{B} \cdot x(t)\\
    y(t) &= \overline{C} \cdot h(t)
\end{aligned}
\label{eq8}
\end{equation}

In addition, for an input sequence of size $T$, a global convolution with kernel $\overline{K}$ can be applied to compute the output of Equation \eqref{eq8} as follows:
\begin{equation}
\begin{aligned}
    \overline{K} &= (C \cdot \overline{B}, C \cdot \overline{AB},..., C \cdot \overline{A^{T-1}B})\\
    y &= x * \overline{K}
\end{aligned}
\label{eq9}
\end{equation}

By introducing the selective scan, the algorithm is further extended to allow the model parameters $B$, $C$ and $\Delta$ to be dynamically tuned to the inputs and to filter out irrelevant information. 

\subsection{Detection Head}
In seminal works \cite{second}, \cite{pointpillars}, anchor-based detection heads are used to predefine axis-aligned anchor points to the head at each location on the input feature map. In contrast, CenterPoint \cite{centerpoint} represents each object by its center point and predicts centerness heat maps where the regression of the bounding box is achieved at each center position. Due to its simplicity and superior performance, we employ center-based detection head in our networks.

\begin{table*}[!t]
\tiny
  \caption{Comparing with the state-of-the-art on DAIR-V2X-I validation dataset in 3D object detection task.
  }
  \centering
  \label{table1}
  \begin{adjustbox}{width=.99\textwidth} 
    \begin{tabular}{l|c|c c c|c c c|c c c}
        
    \toprule
    \multirow{2}{*}{Method} & \multirow{2}{*}{Representation} & \multicolumn{3}{c}{\multirow{1}{*}{Vehicle \textit{(IoU=0.5)}}} & \multicolumn{3}{|c|}{\multirow{1}{*}{Pedestrian \textit{(IoU=0.25)}}} & \multicolumn{3}{c}{\multirow{1}{*}{Cyclist \textit{(IoU=0.25)}}}\\
    \cmidrule{3-11}
     && \multirow{1}{*}{Easy} & \multirow{1}{*}{Mid} & \multirow{1}{*}{Hard} & \multirow{1}{*}{Easy} & \multirow{1}{*}{Mid} & \multirow{1}{*}{Hard} & \multirow{1}{*}{Easy} & \multirow{1}{*}{Mid} & \multirow{1}{*}{Hard}
    \\
    \midrule
    PointRCNN \cite{pointrcnn}&Point&79.20&66.74&66.76&38.85&36.61&36.53&33.47&37.40&37.31\\
    PV-RCNN \cite{pvrcnn}&Point+Voxel&\textbf{86.21}&\textbf{71.39}&\textbf{71.42}&76.34&72.78&72.75&75.40&68.72&69.03\\
    SECOND \cite{second}&Voxel&71.47&53.99&54.00&55.16&52.49&52.52&54.68&31.05&31.19\\
    VoxelRCNN \cite{voxelrcnn}&Voxel&86.18&71.36&71.39&75.53&72.91&72.88&73.41&68.32&68.57\\
    VoxelNeXt \cite{voxelnext}&Voxel&85.19&70.50&70.53&76.35&72.18&72.94&76.84&74.35&74.64\\
    PointPillars \cite{pointpillars}&Pillar&63.07&54.00&54.01&38.53&37.20&37.28&38.46&22.60&22.49\\
    SST \cite{sst}&Pillar&79.55&67.06&67.10&72.41&68.76&68.81&70.85&71.52&71.48\\
    CenterPoint-P \cite{centerpoint}&Pillar&85.07&70.40&70.44&74.59&71.29&71.39&73.47&74.06&74.32\\
    PillarNet \cite{pillarnet}&Pillar&85.04&70.38&70.42&75.34&72.67&72.70&75.92&74.93&75.26\\
    PillarNeXt \cite{pillarnext}&Pillar&84.91&70.23&70.26&75.95&72.56&72.68&75.43&75.30&75.52\\
    \rowcolor{gray!20}
    PillarMamba (ours)&Pillar&85.81&71.04&71.06&\textbf{77.04}&\textbf{73.53}&\textbf{73.54}&\textbf{77.22}&\textbf{75.46}&\textbf{75.73}\\
    \bottomrule 
    \end{tabular}
    \end{adjustbox}
\end{table*}

\begin{figure*}[!t]
\centering
\includegraphics[width=.99\textwidth]{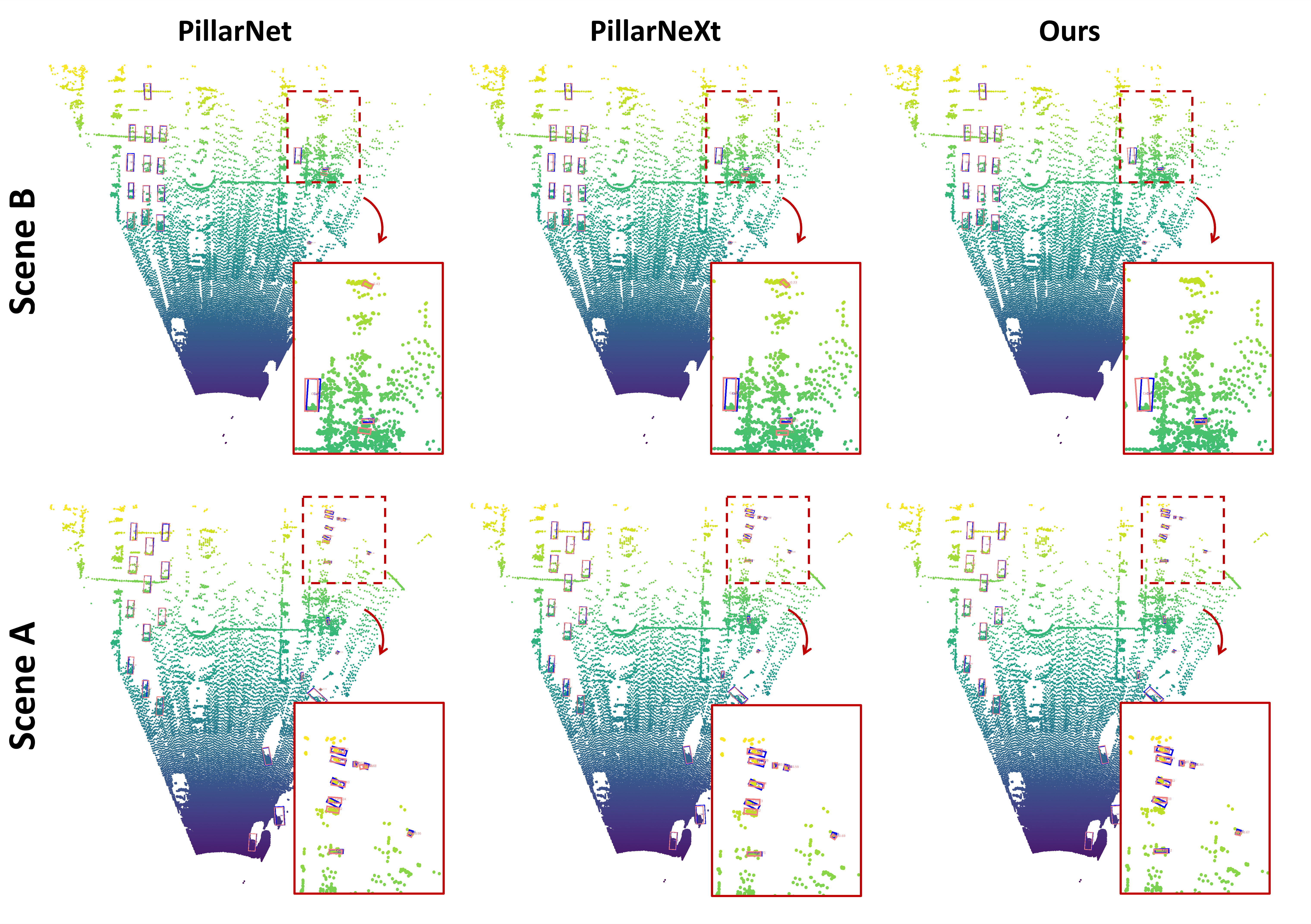}
\caption{We visualize some scenes in BEV map to qualitatively analysis the comparison between our proposed PillarMamba and the state-of-the-art methods (e.g. PillarNet and PillarNeXt). The color distribution of the point cloud is related to the x-axis coordinates, with \textcolor{blue}{blue} bounding boxes that denote the ground truth and \textcolor{red}{red} bounding boxes that denote the predictions of network. It can be seen that, in both scene A and B, PillarNet and PillarNeXt produce incorrect predictions due to the noisy and sparse point cloud at long distances. However, our proposed PillarMamba makes correct predictions in the detection of long-distance in these scenes and correctly classifies sparse point cloud with ambiguous shapes.}
\label{fig_3}
\vspace{-0.2cm}
\end{figure*}

\section{Experiment}
In this section, experiment settings are introduced. Then, the comparison between PillarMamba and state-of-the-art methods is given. Finally, the full-scale experiment to validate the effectiveness of the proposed modules will be presented in detail.

\subsection{Dataset}
\textbf{DAIR-V2X} dataset \cite{dair} introduced a large-scale multi-modal benchmark. The original dataset contains images and point cloud from both vehicle-side and roadside scenes. Specifically, DAIR-V2X-I contains about 10k images and point cloud, of which 50$\%$, 20$\%$ and 30$\%$ of the samples are divided into training, validation and test sets, respectively. However, the test set has not been published so far, so we evaluate the results on the validation set and follow the KITTI evaluation metrics.

\textbf{Metrics.} For DAIR-V2X-I dataset, we report the 40-point average precision (AP$_{3D|R40}$) \cite{ap} of 3D bounding boxes, which is further categorized into three modes: $Easy$, $Middle$ and $Hard$, based on the characteristics of boxes. It includes size, occlusion and truncation, following the metrics of KITTI \cite{kitti}.

\subsection{Experimental Settings}
 We implement our networks in PyTorch and train each model in 80 epochs based on the optimizer AdamW \cite{adamw} and one-cycle schedule \cite{onecycle}. We adopt the widely used data augmentation strategies during training, including the random flip, random rotation, random scaling, and random translation. All training experiments are conducted on 2 RTX-4090 GPUs, and all inference experiments are conducted on 1 RTX-4090 GPU with batch size 1. For DAIR-V2X-I dataset, the detection range is set to [0.0m, 102.4m] horizontally and [-5.0m, 5.0m] vertically, the pillar size is set to 0.2m in x/y-axis.

\subsection{Overall Results}
\textbf{Evaluation on DAIR-V2X-I val set.}
We compare the proposed PillarMamba with state-of-the-art methods on DAIR-V2X-I val set. As shown in Table \ref{table1}, PillarMamba outperforms the state-of-the-art pillar-based methods, demonstrating its effectiveness. Specifically, PillarMamba outperforms the PillarNet \cite{pillarnet} and PillarNeXt \cite{pillarnext} by (0.77, 1.70, 1.30) and (0.90, 1.09, 1.79) AP in vehicle, pedestrian, and cyclist categories, respectively. It addition, although our method loses height information due to the pillar representation, it utilizes a global receptive field to make the pillar-based methods with outperform the voxel-based methods.

\textbf{Visualization results.}
We visualize some scenes in BEV map to qualitatively analysis the comparison between our proposed PillarMamba and the state-of-the-art methods (e.g. PillarNet and PillarNeXt), as shown in Figure \ref{fig_3}. The color distribution of the point cloud is related to the x-axis coordinates, with \textcolor{blue}{blue} bounding boxes that denote the ground truth and \textcolor{red}{red} bounding boxes that denote the predictions of network. It can be seen that, in both scene A and B, PillarNet and PillarNeXt produce incorrect predictions due to the noisy and sparse point cloud at long distances. However, our proposed PillarMamba makes correct predictions in the detection of long-distance in these scenes and correctly classifies sparse point cloud with ambiguous shapes.

\begin{table*}[ht]
\tiny
  \caption{Analysis on the proposed Cross-stage State-space Group. The comparison is the difference between PillarMamba without and with CSG.}
  \centering
  \label{table2}
  \begin{adjustbox}{width=.99\textwidth} 
    \begin{tabular}{c|c c c|c c c|c c c|c|c}

    \midrule
    \multirow{2}{*}{CSG} & \multicolumn{3}{c}{\multirow{1}{*}{Vehicle \textit{(IoU=0.5)}}} & \multicolumn{3}{|c|}{\multirow{1}{*}{Pedestrian \textit{(IoU=0.25)}}} & \multicolumn{3}{c|}{\multirow{1}{*}{Cyclist \textit{(IoU=0.25)}}}
    &\multirow{2}{*}{Latency(ms)}&\multirow{2}{*}{FPS}\\
    \cmidrule{2-10}
    & \multirow{1}{*}{Easy} & \multirow{1}{*}{Mid} & \multirow{1}{*}{Hard} & \multirow{1}{*}{Easy} & \multirow{1}{*}{Mid} & \multirow{1}{*}{Hard} & \multirow{1}{*}{Easy} & \multirow{1}{*}{Mid} & \multirow{1}{*}{Hard}&&
    \\
    \midrule
    -&85.52&70.77&70.79&75.74&72.15&72.24&75.37&73.84&74.04&92.83&10.77\\
    \rowcolor{gray!20}
    \checkmark&\textbf{85.81}&\textbf{71.04}&\textbf{71.06}&\textbf{77.04}&\textbf{73.53}&\textbf{73.54}&\textbf{77.22}&\textbf{75.46}&\textbf{75.73}&\textbf{70.04}&\textbf{14.27}\\
    
    \midrule
    
    \end{tabular}
    \end{adjustbox}
\end{table*}

\begin{table*}[ht]
\tiny
  \caption{Analysis on the proposed Hybrid State-space Block. $LC$ and $RA$ denote the local convolution and residual attention. $Res$ and $Attn$ denote the residual connection and attention module. The vanilla model denotes the vanilla SS2D block with CSG.}
  \centering
  \label{table3}
  \begin{adjustbox}{width=.99\textwidth} 
    \begin{tabular}{c c c|c c c|c c c|c c c}
        
    \midrule
    \multirow{2}{*}{LC} &  \multicolumn{2}{|c}{RA} & \multicolumn{3}{|c}{\multirow{1}{*}{Vehicle \textit{(IoU=0.5)}}} & \multicolumn{3}{|c|}{\multirow{1}{*}{Pedestrian \textit{(IoU=0.25)}}} & \multicolumn{3}{c}{\multirow{1}{*}{Cyclist \textit{(IoU=0.25)}}}\\
    \cmidrule{2-12}
     & \multicolumn{1}{|c}{Res} & Attn & \multirow{1}{*}{Easy} & \multirow{1}{*}{Mid} & \multirow{1}{*}{Hard} & \multirow{1}{*}{Easy} & \multirow{1}{*}{Mid} & \multirow{1}{*}{Hard} & \multirow{1}{*}{Easy} & \multirow{1}{*}{Mid} & \multirow{1}{*}{Hard}
    \\
    \midrule
    -&-&-&85.33&70.62&70.64&74.40&71.45&71.47&73.81&72.82&73.02\\
    \checkmark&-&-&85.45&70.71&70.74&75.30&72.58&72.64&74.88&74.02&74.19\\ 
    \checkmark&\checkmark&-&85.52&70.79&70.80&75.45&72.06&72.12&75.94&74.43&74.63\\
    \rowcolor{gray!20}
    \checkmark&\checkmark&\checkmark&\textbf{85.81}&\textbf{71.04}&\textbf{71.06}&\textbf{77.04}&\textbf{73.53}&\textbf{73.54}&\textbf{77.22}&\textbf{75.46}&\textbf{75.73}\\
    \midrule
    
    \end{tabular}
    \end{adjustbox}
\end{table*}

\begin{table}[ht]
\large
\renewcommand{\arraystretch}{1.2}
\setlength{\heavyrulewidth}{1.2pt}
  \caption{Latency Comparison between pillar-based methods. $Veh$, $Ped$ and $Cyc$ denote AP from vehicle, pedestrian and cyclist at easy mode. $SS$ denotes the single stride backbone.}
  \centering
  \label{table4}
  \begin{adjustbox}{width=.5\textwidth} 
    \begin{tabular}{l|ccc|c|c}  
    \toprule
    \multirow{1}{*}{Method} &
    \multirow{1}{*}{Veh$_\textit{(IoU=0.5)}$} & \multirow{1}{*}{Ped$_\textit{(IoU=0.25)}$} & \multirow{1}{*}{Cyc$_\textit{(IoU=0.25)}$} &
    \multirow{1}{*}{Latency(ms)}&
    \multirow{1}{*}{FPS}
    \\
    \midrule
    SST \cite{sst}&79.55&72.41&70.85&132.70&7.53\\
    PillarNet \cite{pillarnet}&85.04&75.34&75.92&57.16&17.49\\
    PillarNeXt \cite{pillarnext}&84.91&75.95&75.43&\textbf{52.53}&\textbf{19.03}\\
    CenterPoint-SS \cite{centerpoint}&85.73&75.39&76.37&78.84&12.68\\
    \midrule
    \rowcolor{gray!20}
    PillarMamba (ours)&\textbf{85.81}&\textbf{77.04}&\textbf{77.22}&70.04&14.27\\
    \bottomrule 
    \end{tabular}
    \end{adjustbox}
\end{table}

\subsection{Ablation Study}
\textbf{Analysis on the proposed Cross-stage State-space Group.}
In Table \ref{table2}, although the state space model accomplishes efficient feature extraction with its computational burden that expands linearly with sequence length and hardware-accelerated parallel computation, it is still unacceptable when faced with dense contexts. Therefore, CSG is proposed for more efficient computation. It can be seen that it not only achieves computational efficiency and thus inference acceleration, but also allows different layers to complement each other through cross stage connections and enhances the expressive ability of networks.

\textbf{Analysis on the proposed Hybrid State-space Block.}
As shown in Table \ref{table3}, we show step-by-step the performance improvements brought by our proposed components. The proposed HSB effectively solved the local connection disrupted and historical relationship forgotten of standard Mamba thanks to the local convolution and residual attention. The experimental results show that HSB can enhance the network's spatial understanding and preserve the historical memory in recursive state space equation. 

\subsection{Latency}
We performed performance and latency comparison experiments of our proposed PillarMamba with the state-of-the-art pillar-based methods, as shown in Table \ref{table4}. As can be seen, PillarMamba outperforms the the state-of-the-art methods, which demonstrates the effectiveness of the proposed components. For latency, even though PillarMamba is slightly slower than the methods based on sparse backbone, it is still competitive and potential. On the one hand, the sparse convolutional operators  are not easy to deploy and accelerate relative to the dense convolutional operators in the edge scene like roadside scenarios. On the other hand, our PillarMamba achieves advantages in both performance and latency in comparison of dense methods.

\section{Conclusion}
As an important component of Intelligent Transportation Systems (ITS), roadside point cloud 3D object detection has not received sufficient attention and effective network architecture design. The dense context in roadside point cloud due to the perspective difference in sensor locations motivates us to explore the potential of state space model for roadside scenarios. Further, we address the inefficient computation, local connection disrupted and historical relationship forgotten faced by standard Mamba through an effective network structure design. Both quantitative and qualitative results show that our proposed components effectively solve the above challenges. We hope our work can shed light on effective roadside point cloud detectors.

\section*{Acknowledgments}
This work was partially supported by the Key-Area Research and Development Program of Guangdong Province (2023B0909040001).

\bibliographystyle{IEEEtran}
\bibliography{ref}

\begin{IEEEbiographynophoto}{Zhang Zhang} received the B.S. degree in mechanical engineering from Beijing Institute of Technology, Beijing, China, in 2022. He is currently pursuing the Ph.D. degree in mechanical engineering at the National Engineering Laboratory for Electric Vehicles, Beijing Institute of Technology. His research interests include roadside perception and connected vehicles.
\end{IEEEbiographynophoto}
\begin{IEEEbiographynophoto}{Chao Sun} received the B.S. and Ph.D. degree in Mechanical Engineering from Beihang University and Beijing Institute of Technology in 2010 and 2016, respectively. He was a postdoctoral researcher at the Energy, Controls, and Applications Lab in University of California, Berkeley, CA, USA. Currently, he is an Associate Professor at Beijing Institute of Technology, studying on automated and connected vehicles and hybrid electric vehicles.
\end{IEEEbiographynophoto}
\begin{IEEEbiographynophoto}{Chao Yue} received the B.S. degree in vehicle engineering from Beijing Institute of Technology, Beijing, China, in 2023. He is currently working toward a Ph.D. degree in School of Mechanical Engineering, Beijing Institute of Technology, Beijing, China. His current research interests include object detection and cooperative perception for robotic systems and automated vehicles.
\end{IEEEbiographynophoto}
\begin{IEEEbiographynophoto}{Da Wen} received the B.S. degree in vehicle engineering from Beijing Institute of Technology, Beijing, China, in 2021. He is currently pursuing the Ph.D. degree in mechanical engineering at the National Engineering Laboratory for Electric Vehicles, Beijing Institute of Technology. His research interests include decision making in automated vehicles and explainability of artificial intelligent system.
\end{IEEEbiographynophoto}

\begin{IEEEbiographynophoto}{Tianze Wang} received the B.S. degree in vehicle engineering from Beijing Institute of Technology, Beijing, China, in 2023. He is currently pursuing a master's degree at Beijing Institute of Technology, China. His research interests include computer vision applications in vehicle-road fusion perception and driver assistance.
\end{IEEEbiographynophoto}

\begin{IEEEbiographynophoto}{Jianghao Leng} received the B.S. degree in vehicle engineering from Beijing Institute of Technology, Beijing, China, in 2019. He is currently pursuing the Ph.D. degree in mechanical engineering at the National Engineering Laboratory for Electric Vehicles, Beijing Institute of Technology. From 2023 to 2024, he was a visiting student at the College of Design and Engineering, National University of Singapore. His research interests include multi-sensor fusion SLAM and Eco-driving for autonomous vehicles. 
\end{IEEEbiographynophoto}

\vfill
\end{document}